\documentclass[runningheads]{llncs}

\usepackage{eccv}

\usepackage{multirow}
\usepackage[dvipsnames,table]{xcolor}
\definecolor{mygray}{gray}{.9}
\definecolor{ggray}{RGB}{127,127,127}
\definecolor{reda}{RGB}{192,0,0}
\definecolor{redb}{RGB}{217,148,143}
\definecolor{myyellow}{RGB}{190,144,0}
\definecolor{mygreen}{RGB}{80,100,40}
\definecolor{myblue}{RGB}{30,90,100}
\definecolor{dark-gray}{gray}{0.20}
\definecolor{middle-gray}{gray}{0.85}
\definecolor{light-gray}{gray}{0.93}
\definecolor{lightblue}{rgb}{0.85, 0.95, 1}
\definecolor{lighterblue}{rgb}{0.9, 0.97, 1}
\definecolor{palestblue}{rgb}{0.95, 0.98, 1}

\usepackage{eccvabbrv}

\usepackage{graphicx}
\usepackage{booktabs}
\usepackage{caption} 

\usepackage[accsupp]{axessibility}  

\usepackage{hyperref}

\usepackage{orcidlink}

\begin{document}

\title{TimeThink: Reasoning with Time for Video LLMs} 


\author{
Handong Li\inst{1,2}\thanks{Equal contribution.}
\and
Longteng Guo\inst{2}$^{\star}$
\and
Zikang Liu\inst{1,2}$^{\star}$
\and
Dongze Hao\inst{3}
\and
Yepeng Tang\inst{2}
\and
Zijia Zhao\inst{2}
\and
Jie Jiang\inst{2}
\and
Zhiwei Jin\inst{3}
\and
Chen Chen\inst{3}
\and
Haonan Lu\inst{3}
\and
Jing Liu\inst{1,2}\thanks{Corresponding author.}
}

\authorrunning{H.~Li et al.}

\institute{
School of Artificial Intelligence, University of Chinese Academy of Sciences
\and
Institute of Automation, Chinese Academy of Sciences
\and
OPPO AI Center, OPPO Inc.
\\
\email{\{lihandong2023\}@ia.ac.cn, \{longteng.guo,jliu\}@nlpr.ia.ac.cn}}

\makeatletter
\let\savedtitle\@title
\makeatother
\maketitle

\begin{abstract}
Video reasoning requires models to identify and verify temporally localized evidence within long video sequences. 
Recent Video Large Language Models (Video-LLMs) have shown promising reasoning abilities when aligned with reinforcement learning, yet existing approaches typically rely on outcome-based rewards that supervise only the final prediction. 
Such supervision provides limited guidance on how models should discover the relevant temporal evidence during intermediate reasoning.
In this work, we propose \textbf{TimeThink}, a reinforcement learning framework that explicitly guides temporal evidence discovery in Video-LLMs. 
Our key idea is to treat temporal clue steps as the fundamental optimization primitive of video reasoning, where each reasoning step references a candidate time interval in the video. 
We introduce a step-wise temporal process reward that provides localized credit assignment for these clues and a joint process--outcome optimization objective that balances reasoning fidelity with task correctness. 
To enable scalable training, we construct TimeThink-RFT-20K, a dataset with automatically derived temporal evidence segments.
Extensive experiments across video reasoning, temporal grounding, and general video understanding benchmarks show that TimeThink consistently improves both temporal localization and reasoning performance, achieving state-of-the-art results among open-source video RL models.

\keywords{Video-LLMs \and Reinforcement Learning \and Process Reward}

\end{abstract}    
\section{Introduction}
\label{sec:intro}

\begin{figure}[th]
    \centering
    \includegraphics[width=1.0\linewidth]{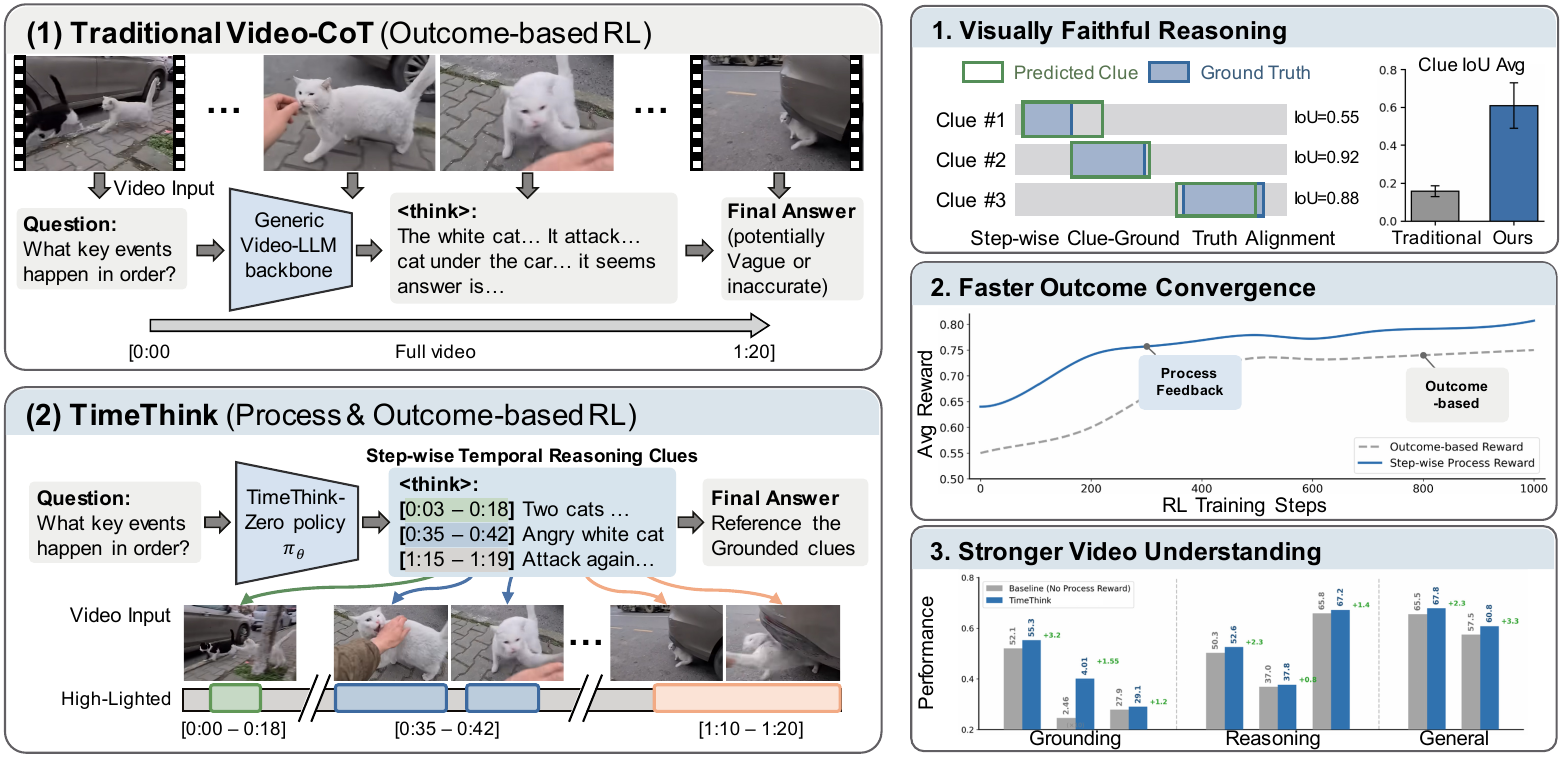}
    \caption{\textbf{TimeThink vs. Traditional Video-CoT.} 
    \textbf{(Left)} Traditional Video-CoT relies on outcome-based rewards that supervise only the final answer, providing limited guidance for temporal evidence discovery. 
    TimeThink instead models reasoning as a sequence of \emph{temporal clue steps}, where each step references a candidate time interval in the video. 
    \textbf{(Right)} Step-wise temporal process rewards supervise these clues, enabling (1) temporally grounded reasoning, (2) faster RL convergence, and (3) stronger video understanding.}
    \label{fig:pre_exp}
\end{figure}

Recent advances in Video Large Language Models (Video-LLMs)~\cite{Qwen2VL,Qwen2_5VL,zhang2024video,li2025breaking} have significantly expanded the frontier of machine video understanding.
By integrating pretrained vision encoders with large language models, these systems are capable of performing diverse tasks such as video question answering, event narration, and complex video reasoning~\cite{li2024mvbench,chen2023vast,zhaoneedle,zhou2024mlvu,wu2024longvideobench}.
However, enabling models to reason reliably over long temporal contexts remains a central challenge.
Unlike static images, videos encode dynamic processes where events unfold over time and critical evidence may appear only within brief segments of a long sequence.
Consequently, effective video reasoning requires models to progressively inspect and verify temporally localized evidence rather than relying solely on global semantic impressions of the video.

To improve reasoning capabilities, recent work has begun to apply reinforcement learning (RL) to Video-LLMs~\cite{feng2025video,zhao2025r1,wang2025time,li2025videochat}.
Algorithms such as Group Relative Policy Optimization (GRPO)~\cite{shao2024deepseekmath} enable models to explore multiple reasoning trajectories and optimize them using reward signals.
In these frameworks, the model generates intermediate reasoning steps before producing a final answer, and the learning objective encourages trajectories that lead to correct predictions.
While such approaches have shown promising improvements in reasoning ability, existing RL-based methods primarily rely on \emph{outcome-based rewards} that evaluate only the final prediction, such as answer correctness or task-specific localization metrics.

To address this limitation, we propose to explicitly guide the discovery of temporal evidence during reasoning.
We observe that many video reasoning tasks can be naturally decomposed into a sequence of localized inspections along the timeline, where each step focuses on verifying whether a short temporal interval contains evidence relevant to the query.
Rather than requiring explicit supervision of the entire reasoning trajectory, which is difficult to obtain at scale, we instead guide the reasoning process through lightweight temporal references that indicate where the model should inspect the video.

Based on this insight, we introduce \textbf{TimeThink}, a reinforcement learning framework that learns temporally grounded reasoning in Video-LLMs.
The central idea of TimeThink is to treat \emph{temporal clue steps} as the fundamental optimization primitive for video reasoning.
Instead of generating an unconstrained reasoning trajectory, the model is encouraged to iteratively reference short temporal intervals in the video and verify whether they contain evidence relevant to the query.
Each reasoning step therefore corresponds to inspecting a candidate time span that may support or refute the current hypothesis.

To guide this behavior, we design a \emph{step-wise temporal process reward} that evaluates the alignment between the referenced interval and the evidence segments associated with the query.
This reward provides localized credit assignment for individual reasoning steps, encouraging the model to progressively identify informative temporal regions while reasoning.
Compared with outcome-only rewards that supervise only the final prediction, this process-level feedback directly promotes temporally grounded reasoning and reduces reliance on coarse global correlations.

In addition to the process reward, we introduce a \emph{joint process–outcome optimization} objective that balances temporal reasoning quality with task-level correctness.
The process reward encourages accurate discovery of temporal evidence during intermediate reasoning, while the outcome reward ensures that the final response remains aligned with the target task objective.
Together, these signals enable the model to learn reasoning trajectories that are both temporally faithful and task-effective.

To support scalable process reward computation, we construct \textbf{TimeThink\-RFT-20K}, a dataset of video question--answer pairs augmented with automatically derived temporal evidence segments.
These evidence segments provide supervision signals for evaluating temporal clue steps during RL training.

We evaluate TimeThink across a diverse set of video reasoning, general video understanding, and temporal grounding benchmarks.
TimeThink consistently improves both temporal localization accuracy and reasoning performance.
Across seven benchmarks, TimeThink achieves state-of-the-art performance among open-source video RL models and consistently outperforms baselines restricted solely to outcome rewards, demonstrating the effectiveness of process-level temporal supervision.

Our contributions are summarized as follows:
\begin{itemize}
\item We introduce temporal clue steps as a reasoning primitive for video LLMs, modeling video reasoning as a sequence of temporally grounded evidence inspections.

\item We propose TimeThink, a reinforcement learning framework that employs a step-wise temporal process reward and a joint process--outcome optimization objective to encourage temporally grounded reasoning.

\item Extensive experiments across multiple video reasoning, grounding, and understanding benchmarks demonstrate that TimeThink significantly improves both temporal localization and general video reasoning performance.

\item We will release the code, models, and the dataset to facilitate reproducible research on temporally grounded video reasoning.
\end{itemize}

\section{Related Work}

\subsection{Video Large Language Models}
Recent advancements in Large Language Models (LLMs) have driven the rapid development of Video Large Language Models (Video-LLMs)~\cite{li2024mvbench,cheng2024videollama,instructblip,maaz2023video,li2025llama}. The prevailing alignment paradigm heavily relies on Supervised Fine-Tuning (SFT)~\cite{lin2023video,xu2024pllava,ataallah2024minigpt4,zhang2024explore}. While SFT establishes strong zero-shot capabilities on general video question-answering~\cite{msvd,yu2019activitynet,maaz2023video,mangalam2023egoschema,li2024mvbench,fu2024video}, it inherently struggles with complex temporal reasoning and precise localization, often overfitting to linguistic priors and visual biases. To overcome these limitations, recent Video-LLM research~\cite{feng2025video,li2025videochat,li2025tempsamp} emulates reasoning-centric LLMs~\cite{jaech2024openai,guo2025deepseek,yang2025qwen3} by adopting Reinforcement Learning (RL). Unlike the deterministic constraints of SFT, RL captures comprehensive alignment supervision by empowering models to actively explore diverse reasoning trajectories.

\subsection{Video Reinforcement Fine-Tuning}
Recent Reinforcement Fine-Tuning (RFT) efforts in Video-LLMs largely follow two trajectories. The first focuses on enhancing general multimodal capabilities. Models like Video-R1~\cite{feng2025video} and VideoChat-R1~\cite{li2025videochat} scale RL to the video domain by structuring complex suites of task-specific outcome rewards to optimize diverse skills. However, these outcome-centric approaches require arduous manual design and struggle to scale efficiently across heterogeneous benchmarks. The second trajectory targets the temporal reasoning bottleneck using targeted IoU-based rewards. For instance, TimeZero~\cite{wang2025timezero} utilizes strict Temporal IoU outcome rewards, while TempSamp-R1~\cite{li2025tempsamp} integrates temporal alignment constraints directly into advantage estimation. Despite their efficacy in precise localization, these highly specialized optimizations consistently fail to transfer acquired grounding capabilities to broader, open-ended video understanding tasks.
\section{Video Reasoning with Temporal Clues}
\label{sec:reasoning_clues}

Given a video $V$ and a textual query $x$, a Video-LMM is required to produce a response sequence $y$. 
In our formulation, the response is structured into an intermediate reasoning trajectory and a final answer:
\begin{equation}
y = (y^{think}, y^{ans}) \, ,
\end{equation}
where $y^{think}$ denotes the reasoning process enclosed in a \texttt{<think>} block and $y^{ans}$ represents the final answer enclosed in a \texttt{<answer>} block.

The reasoning trajectory is composed of a sequence of temporally anchored reasoning units
\begin{equation}
y^{think} = (s_1, s_2, ..., s_K) \, ,
\end{equation}
where each $s_k$ denotes a temporal reasoning step that operates on a specific portion of the video, e.g., verifying whether an event occurs, locating when it happens, or ruling out irrelevant intervals.
We associate each step with a temporal clue span
\begin{equation}
p_k = (t_k^{start}, t_k^{end}) \, ,
\end{equation}
which specifies the video interval referenced by that step.

Intuitively, video reasoning unfolds as a sequence of localized inspections over the temporal structure of the video. 
At each step, the model examines a short interval, extracts relevant observations, and incrementally aggregates these clues to derive the final answer. 
A typical reasoning trajectory therefore consists of several temporally grounded observations followed by a concise conclusion.

\begin{figure}[t]
\centering
\includegraphics[width=1.0\linewidth]{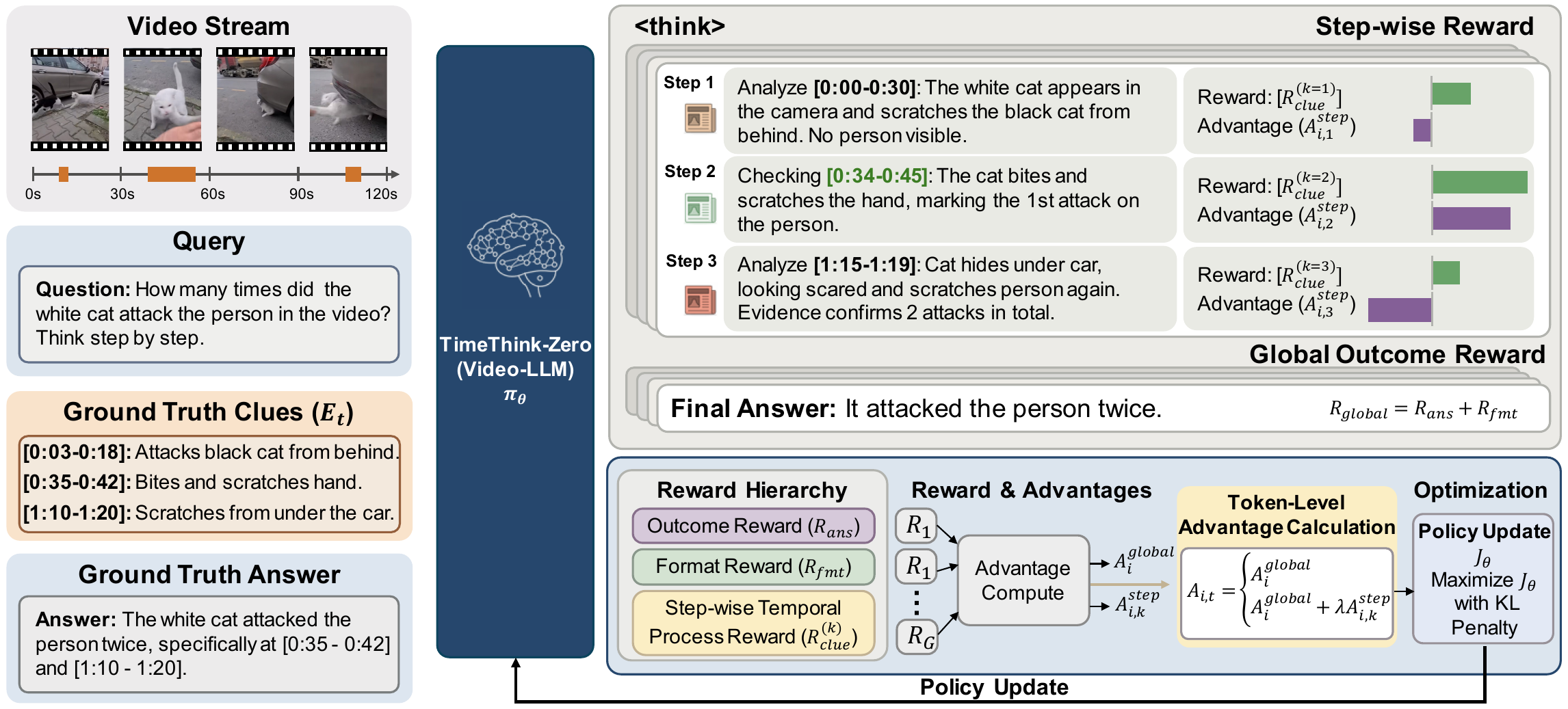}
\caption{\textbf{Architecture of TimeThink.} The framework processes input video and textual queries through a Video-LLM. During the reinforcement learning phase, the model generates an intermediate reasoning trajectory, which is supervised by a step-wise process reward alongside outcome-based rewards to optimize the policy $\pi_\theta$.
}
\label{fig:method}
\end{figure}

\section{TimeThink Framework}
Training Video-LLMs to generate temporally grounded reasoning trajectories is non-trivial. 
Although modern video foundation models possess strong visual and language capabilities, they are typically optimized with outcome-level objectives that evaluate only the final prediction. 
Such supervision provides little guidance on how the model should identify and verify relevant temporal evidence during intermediate reasoning.

To address this limitation, we introduce \textbf{TimeThink}, a reinforcement learning framework for learning temporally grounded reasoning in Video-LLMs. 
Built on top of a pre-trained Video-LMM, TimeThink optimizes the reasoning trajectory using reward signals that evaluate both the final prediction and the temporal evidence referenced during reasoning. 
Instead of supervising the entire reasoning trace explicitly, the model is encouraged to discover grounded reasoning behaviors through reinforcement learning.

\subsection{Temporal Clue Step as the Optimization Primitive}
A key design choice in TimeThink is to treat temporally grounded reasoning steps as fundamental optimization units. As introduced in \cref{sec:reasoning_clues}, the reasoning trajectory constitutes a sequence of temporal clue steps, where each step inspects specific video intervals and extracts observations relevant to the query. Rather than evaluating the reasoning trajectory only at the sequence level, TimeThink attaches learning signals directly to these clue steps. 
This design enables localized credit assignment during training and encourages the model to progressively identify informative temporal regions while reasoning.

Based on this formulation, we introduce a step-wise temporal process reward that evaluates the quality of temporal clues referenced during reasoning.

\subsection{Step-wise Temporal Process Reward}

Given the temporal clue step abstraction, the next question is how to evaluate the quality of these steps during training. 
TimeThink introduces a step-wise temporal process reward that measures whether the temporal interval referenced in a reasoning step corresponds to relevant visual evidence.

Let $E_{gt}$ denote the set of ground-truth temporal segments associated with the query. 
For each reasoning step that references a temporal clue $p_k$, we compute the reward based on the temporal overlap between the referenced interval and the ground-truth evidence:
\begin{equation}
R_{clue}^{(k)} = \max_{g \in E_{gt}} \text{IoU}(p_k, g),
\end{equation}
where $\text{IoU}(\cdot)$ denotes temporal intersection-over-union between two intervals.

This reward evaluates whether the temporal interval inspected at a reasoning step aligns with relevant visual evidence. 
Steps referencing intervals that overlap with ground-truth segments receive higher rewards, while inspections of irrelevant regions yield low or zero reward.

The IoU formulation naturally balances temporal coverage and localization precision. Consequently, the model cannot trivially maximize the reward by expanding the predicted interval, since overly long spans reduce the overlap ratio.

Importantly, the reward is computed based on the temporal interval referenced in each reasoning step rather than the textual content itself. 
This design provides fine-grained credit assignment during training while allowing the model to flexibly explore intermediate reasoning trajectories in natural language.

\subsection{Joint Process--Outcome Optimization}

The temporal process reward supervises whether the model inspects relevant evidence during reasoning, while the final response must still satisfy outcome-level objectives such as answer correctness and output format. 
Since these signals differ in both granularity and scale, TimeThink combines them through a token-level advantage construction under the GRPO training paradigm.

Given an input prompt $x$, we sample a group of $G$ responses $\{y_i\}_{i=1}^{G}$. 
For each response, we compute a global outcome reward
$R_{global}^{(i)} = R_{ans}^{(i)} + R_{fmt}^{(i)}$,
where $R_{ans}$ evaluates the final answer correctness (e.g., exact match or task-specific metrics) and $R_{fmt}$ encourages adherence to the required output structure (e.g., proper \texttt{<think>} and \texttt{<answer>} tags).
We then form the standardized global advantage by group-wise normalization:
\begin{equation}
A_i^{global} =
\frac{R_{global}^{(i)} - \mu_{global}}
{\sigma_{global} + \epsilon},
\end{equation}
where $\mu_{global}$ and $\sigma_{global}$ are computed over the $G$ sampled responses.

In parallel, for each response $y_i$, we compute step rewards $\{R_{clue}^{(k)}\}$ for all temporal clue steps. 
Because the number of steps varies across responses, we aggregate all valid step rewards across the group and normalize them to obtain step-level advantages:
\begin{equation}
A_{i,k}^{step} =
\frac{R_{clue}^{(k)} - \mu_{step}}
{\sigma_{step} + \epsilon},
\end{equation}
where $\mu_{step}$ and $\sigma_{step}$ are computed over the flattened set of step rewards.

We construct a token-wise advantage that aligns with the two-block output structure. 
Tokens in the \texttt{<think>} block are additionally shaped by the step-level grounding signal, while tokens in the \texttt{<answer>} block are optimized only with the global outcome signal. 
Let $k(t)$ denote the index of the temporal clue step that token $t$ belongs to within $y_i^{think}$. We define
\begin{equation}
A_{i,t} =
\begin{cases}
A_i^{global} + \lambda A_{i,k(t)}^{step}, & t \in y_i^{think}, \\
A_i^{global}, & t \in y_i^{ans}.
\end{cases}
\end{equation}
Here, $\lambda$ controls the strength of process supervision, with a default value of 0.5.

The policy parameters are then optimized under the GRPO objective with a KL regularization term that constrains the policy from drifting too far from a fixed reference model $\pi_{ref}$:
\begin{equation}
J(\theta) =
\mathbb{E}
\left[
\frac{1}{G}
\sum_{i=1}^{G}
\sum_{t=1}^{|y_i|}
\left(
A_{i,t}\log \pi_\theta(y_{i,t}|x,y_{i,<t})
-
\beta D_{KL}(\pi_\theta || \pi_{ref})
\right)
\right].
\end{equation}
Through this joint optimization, TimeThink improves answer correctness and encourages temporally grounded reasoning trajectories, enabling the model to progressively identify and verify relevant visual evidence during inference.

\subsection{Two-Stage Reinforcement Fine-Tuning}

We adopt a two-stage reinforcement fine-tuning (RFT) strategy to train TimeThink, which first establishes temporally grounded reasoning behavior and then generalizes it to broader video understanding tasks.

{\textbf{Stage 1: Temporal-Grounded RFT.}}
We first train the model on the Time\-Think-RFT-20K dataset (introduced in Sec.~\ref{sec:data}) to encourage temporally grounded reasoning. 
During this stage, the model is encouraged to reference temporal clues when generating intermediate reasoning within the \texttt{<think>} block. 
Training is guided by the temporal process reward $R_{clue}$ together with outcome rewards for answer correctness and output format. 
This stage encourages the model to inspect relevant video intervals and ground its reasoning in temporal evidence.

{\textbf{Stage 2: Generalized Outcome RFT.}}
After the model acquires stable temporal reasoning behavior, we expand the training distribution to include the full LLaVA-Video-178K dataset together with 77K timestamp-grounded samples from DiDeMo~\cite{anne2017localizing} and ActivityNet Captions~\cite{krishna2017dense}. 
In this stage, the strict temporal process reward is relaxed and training primarily relies on outcome-level rewards. 
This allows the model to apply the learned reasoning strategy to a broader range of open-ended video understanding tasks while maintaining flexibility in its reasoning trajectories.

\section{TimeThink-RFT-20K Dataset}
\label{sec:data}

To support process-level supervision for temporally grounded reasoning, we construct the \textbf{TimeThink-RFT-20K} dataset. Each instance contains a video question–answer pair and temporal evidence segments indicating where relevant visual information appears. These annotations are used to compute the step-wise temporal process reward during reinforcement learning.

Each sample is represented as $(Q, A, \{(s_k, e_k)\}_{k=1}^{K})$,
where $Q$ denotes the question, $A$ the answer, and $(s_k, e_k)$ the temporal boundaries of video segments that contain evidence relevant to answering the question. We derive the dataset from LLaVA-Video-178K~\cite{zhang2024video}, which provides diverse video question–answer pairs. For each video, we first segment it into candidate temporal clips using \texttt{PySceneDetect}~\cite{Castellano_PySceneDetect}, which detects shot boundaries and produces fine-grained temporal segments.

We then employ a teacher model, Qwen3-VL-235B~\cite{bai2025qwen3}, to estimate the semantic relevance between each candidate clip and the corresponding question–answer pair.
The teacher assigns a relevance score in the range $[0,10]$, and clips with scores above a predefined threshold are retained as temporal evidence candidates.

After removing videos with excessive fragmentation or ambiguous evidence, we obtain a dataset of approximately 20K samples. 
The resulting annotations provide coarse temporal evidence associated with each Q-A pair. 
These segments are used to compute the temporal process reward during reinforcement learning, which evaluates the overlap between predicted clues and evidence intervals. 
This design allows the model to learn temporally grounded reasoning from approximate evidence without requiring dense manual annotations.

\section{Experiment}~\label{sec:exp}

In this section, we detail the experimental setup and dataset configurations, followed by a comprehensive comparison with state-of-the-art methods across multiple video understanding benchmarks.

{\textbf{Implementation Details.}}
TimeThink builds upon the robust, non-reasoning Qwen2.5-VL-7B~\cite{Qwen2VL} backbone, directly applying Group Relative Policy Optimization (GRPO) without relying on a Supervised Fine-Tuning (SFT) cold-start. Video inputs are sampled at 2 frames per second (fps) to preserve comprehensive temporal dynamics while maintaining the native resolution for optimal spatial perception. Following the foundational backbone architecture, this configuration supports a maximum visual sequence length of 24K tokens and an overall maximum context window of 32K tokens. The policy is optimized using a learning rate of $1\times10^{-6}$, a global batch size of 64, and a GRPO group size of 8. Training is executed via the advanced ms-swift~\cite{zhao2025swift} framework; the first and second stages require one day and three days, respectively, utilizing 32 NVIDIA H100 GPUs. Comprehensive hyperparameter configurations are detailed in the Appendix.

{\textbf{Training Data.}}
Our training paradigm entirely bypasses the conventional SFT cold-start, initiating directly with a two-stage GRPO pipeline. During the first stage, we utilize our curated TimeThink-RFT-20K dataset to enforce rigorous adherence to the prescribed structural format, concurrently training the model to ground its intermediate reasoning trajectory using critical temporal clues. In the subsequent stage, we transition to the LLaVA-Video-178K~\cite{zhang2024video} dataset, which serves as a foundational corpus for generalized video comprehension. Furthermore, we augment this distribution with 77K timestamp-grounded samples from DiDeMo~\cite{anne2017localizing} and ActivityNet Captions~\cite{krishna2017dense} to further enhance the model's proficiency in identifying and extracting key temporal information.

\subsection{Main Results}

We evaluate TimeThink across a comprehensive suite of video-language benchmarks, categorized as follows: 1) Video Reasoning Benchmarks, including VideoMMMU~\cite{hu2025video}, VSIBench~\cite{yang2025thinking}, and MMVU~\cite{zhao2025mmvu}; 2) General Video Understanding, comprising MLVU~\cite{zhou2024mlvu}, VideoMME~\cite{fu2024video}, LongVideoBench~\cite{wu2024longvideobench}, and MVBench~\cite{li2023mvbench}; 3) Temporal Video Benchmarks, utilizing Charades-STA~\cite{gao2017tall}, CGBench~\cite{chen2024cg}, and NExT-GQA~\cite{xiao2024can}. We benchmark the proposed TimeThink framework against state-of-the-art Supervised Fine-Tuning (SFT) and Reinforcement Learning (RL) methodologies, ensuring a rigorous comparison by employing the identical Qwen\-2.5-VL-7B~\cite{Qwen2_5VL} backbone architecture across all baseline models. Furthermore, to guarantee standardized evaluations and ensure full reproducibility, all quantitative assessments are conducted utilizing the unified evaluation protocols provided by the \texttt{lmms-eval} framework~\cite{zhang2024lmmsevalrealitycheckevaluation,lmms_eval2024}.

\subsection{Grounding Task Evaluation}

\begin{table}[t]
\centering

\begin{minipage}[t]{0.48\linewidth}
\caption{\textbf{Zero-Shot Grounded VideoQA on CGBench.} The results in \textbf{bold} indicate the best performance.}
\label{tab:cgbench_single}
\resizebox{\linewidth}{!}{%
\begin{tabular}{lc ccc}
\toprule
Method & Size & mIoU & R@IoU & A@IoU \\
\midrule
\multicolumn{3}{l}{\textbf{\textit{Proprietary Models}}} \\
GPT-4o~\cite{hurst2024gpt} & -- & 5.62 & 8.30 & {4.38} \\
Gemini-1.5-Pro~\cite{team2024gemini} & -- & 3.95 & 5.81 & {2.53} \\
\midrule
\multicolumn{3}{l}{\textbf{\textit{Supervised Fine-Tuning Models}}} \\
Qwen2-VL~\cite{Qwen2VL} & 72B & 3.58 & 5.32 & 2.54 \\
ShareGPT4Video~\cite{chen2024sharegpt4video} & 16B & 1.85 & 2.65 & 1.01 \\
VideoCCAM~\cite{fei2024video} & 14B & 2.63 & 3.53 & 1.76 \\
LLaVA-OV~\cite{liu2024visual} & 13B & 1.63 & 1.78 & 1.01 \\
Videochat2~\cite{li2024mvbench} & 7B & 1.28 & 1.98 & 0.94 \\
MiniCPM-v2.6~\cite{hu2024minicpm} & 8B & 2.35 & 2.96 & 1.35 \\
LongVA~\cite{zhang2024long} & 7B & 2.94 & 3.86 & 1.78 \\
InternVL2~\cite{chen2024far} & 7B & 3.91 & 5.05 & 2.64 \\
{Qwen2.5-VL-SFT} & 7B & { 4.01} & {5.54} & {3.31} \\
\midrule
\multicolumn{3}{l}{\textbf{\textit{Reinforcement Fine-Tuning Models}}} \\
{Qwen2.5-VL-GRPO} & 7B & {4.88 } & { 7.39} & {3.85 } \\
\rowcolor{lightblue} 
\textbf{TimeThink (Ours)} & 7B & \textbf{5.55} & \textbf{8.27} & \textbf{4.01} \\
\bottomrule
\end{tabular}%
}
\end{minipage}\hfill
\begin{minipage}[t]{0.49 \linewidth}
\vspace{0pt}
\caption{\textbf{Zero-Shot Grounded VideoQA on NExT-GQA.} The results in \textbf{bold} indicate the best performance.}
\label{tab:nextgqa_single}
\resizebox{\linewidth}{!}{%
\begin{tabular}{lc ccc}
\toprule
Method & Size & mIoU & mIoP & Acc@GQA \\
\midrule
\multicolumn{3}{l}{\textbf{\textit{Supervised Fine-Tuning Models}}} \\
FrozenBiLM NG+~\cite{yang2022zero} & 890M & 9.6 & 24.2 & 17.5 \\
VIOLETv2~\cite{fu2023empirical} & - & - & 23.6 & 12.8 \\
SeViLA~\cite{yu2023self} & 4B & 21.7 & 29.5 & 16.6 \\
LangRepo~\cite{kahatapitiya2025language} & 8$\times$7B & 18.5 & 27.1 & 14.9 \\
VideoStreaming~\cite{qian2024streaming} & 8.3B & 19.4 & 30.8 & 18.1 \\
LLoVi~\cite{zhang2024simple} & 1.8T & - & 24.3 & 24.3 \\
HawkEye~\cite{wang2024hawkeye} & 7B & 25.7 & 33.4 & 23.5 \\
Grounding-VideoLLM~\cite{wang2024grounded} & 4B & 21.1 & 34.5 & 26.7 \\
{Qwen2.5-VL-SFT} & 7B & {28.6 } & { 35.8} & {27.7 } \\
\midrule
\multicolumn{3}{l}{\textbf{\textit{Reinforcement Fine-Tuning Models}}} \\
VideoChat-TPO~\cite{yan2025task} & 7B & 27.7 & 35.6 & 25.5 \\
VideoMind~\cite{liu2025videomind} & 1.5B & 28.6 & 35.6 & 25.2 \\
VideoMind~\cite{liu2025videomind} & 7B & 31.4 & 39.0 & 28.2 \\
{Qwen2.5-VL-GRPO} & 7B & {34.0 } & {37.4 } & { 28.2} \\
\rowcolor{lightblue}
\textbf{TimeThink (Ours)} & 7B & \textbf{35.8} & \textbf{40.3} & \textbf{29.1} \\
\bottomrule
\end{tabular}%
}\end{minipage}
\end{table}

\begin{table}[t]
\centering

\begin{minipage}[t]{0.5\linewidth}
\caption{\textbf{Zero-Shot Temporal Grounding on Charades-STA.} The results in \textbf{bold} indicate the best performance.}
\label{tab:charades_single}
\resizebox{\linewidth}{!}{%
\begin{tabular}{lcccc}
\toprule
Method & Size & R@0.5 & R@0.7 & mIoU \\
\midrule
\multicolumn{3}{l}{\textbf{\textit{Proprietary Models}}} \\
GPT-4o~\cite{hurst2024gpt} & -- & - & - & 35.7 \\
\midrule
\multicolumn{3}{l}{\textbf{\textit{Supervised Fine-Tuning Models}}} \\
VTimeLLM~\cite{huang2024vtimellm} & 13B & 34.3 & 14.7 & 34.6 \\
Qwen2.5-VL~\cite{Qwen2_5VL} & 72B & - & - & 50.9 \\
Qwen2.5-VL~\cite{Qwen2_5VL} & 7B & - & - & 43.6 \\
HawkEye~\cite{wang2024hawkeye} & 7B & 31.4 & 14.5 & 34.7 \\
ChatVTG~\cite{qu2024chatvtg} & 7B & 33.0 & 15.9 & 34.9 \\
TimeMarker~\cite{chen2024timemarker} & 8B & 51.0 & 26.9 & 48.4 \\
TimeSearch~\cite{pan2025timesearch} & 7B & 52.4 & 24.5 & 48.6 \\
{Qwen2.5-VL-SFT} & 7B & {58.1} & {33.0} & {51.8} \\
\midrule
\multicolumn{3}{l}{\textbf{\textit{Reinforcement Fine-Tuning Models}}} \\
VideoChat-TPO~\cite{yan2025task} & 7B & 40.2 & 20.8 & 38.1 \\
VideoMind~\cite{liu2025videomind} & 7B & 59.1 & 31.2 & 50.2 \\
{Qwen2.5-VL-GRPO} & 7B & {59.4} & {34.4} & {52.4} \\
\rowcolor{lightblue} 
\textbf{TimeThink (Ours)} & 7B & \textbf{64.1} & \textbf{38.1} & \textbf{55.3} \\
\bottomrule
\end{tabular}%
}
\end{minipage}\hfill
\begin{minipage}[t]{0.49\linewidth}
\vspace{0pt}
\centering
\includegraphics[width=\linewidth]{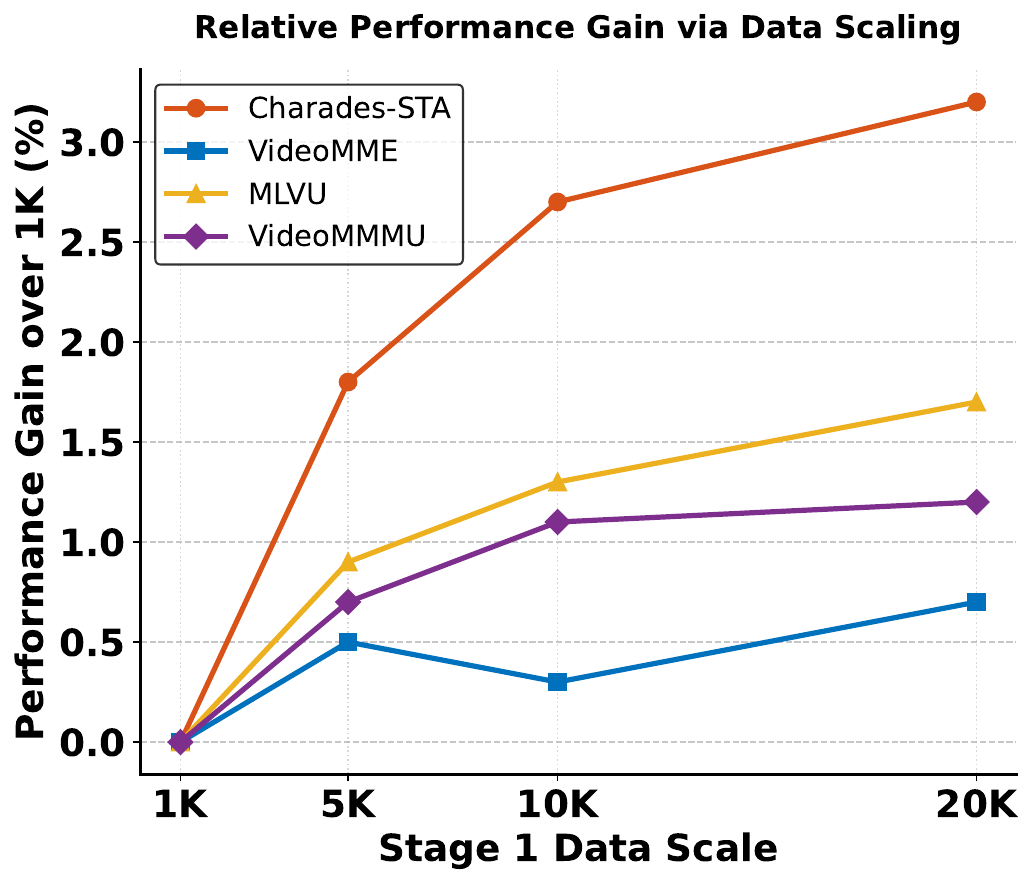}
\captionof{figure}{\textbf{Data Scaling Gains.} Relative performance improvements achieved across multiple benchmarks when scaling the Stage 1 data from 1K to 20K, evaluated after identical Stage 2 training.}
\label{fig:data_scaling}
\end{minipage}

\end{table}
\begin{table*}[t]
\centering
\captionof{table}{\textbf{Performance on zero-shot video-language benchmarks.} We evaluate TimeThink on 3 video reasoning benchmarks and 4 general video benchmarks. The results in \textbf{bold} and \underline{underline} values indicate the best and second-best performance among Reinforcement Fine-Tuning Models, respectively.}
\vspace{-2mm}
\label{tab:general}
\resizebox{1.0\linewidth}{!}{
\begin{tabular}{l c ccc cccc}
\toprule
\multirow{2}{*}{\textbf{Model}} & \multirow{2}{*}{\textbf{Size}}  & \multicolumn{3}{c}{\textbf{Reasoning Tasks}}  & \multicolumn{4}{c}{\textbf{General Tasks}} \\ 
\cmidrule(lr){3-5} \cmidrule(lr){6-9}  
&  & \multirow{1}{*}{\textbf{VideoMMMU}}& \multirow{1}{*}{\textbf{VsiBench}}& \multirow{1}{*}{\textbf{MMVU}} &   \multirow{1}{*}{\textbf{MLVU}} & \multirow{1}{*}{\textbf{VideoMME}} & \multirow{1}{*}{\textbf{LongVideo}}  & \multirow{1}{*}{\textbf{MVBench}}\\ 
\midrule
\multicolumn{3}{l}{\textbf{\textit{Proprietary Models}}} \\
GPT-4V~\cite{OpenAI2023GPT4TR}  & - & -  & - &  -  & 49.2  & 59.5 & 59.1  & 43.7 \\
GPT-4o~\cite{hurst2024gpt}  & -  & 61.2  & 34.0 &  67.4 & 64.6  & 71.9 & 66.7   & -  \\
Gemini-1.5-Pro~\cite{team2024gemini}   & -  &  53.9  & 48.8  &  65.4 & -  & 75.0 & 64.0   & - \\
\midrule
\multicolumn{3}{l}{\textbf{\textit{Supervised Fine-Tuning Models}}} \\
VILA-1.5~\cite{lin2024vila} & 40B & 34.0  & 31.2 &  - & -   & 60.1 & -    & - \\
LLaVA-OV~\cite{liu2024visual}  & 7B  & 33.9  & 32.4 &  54.7 & {63.7}   & 58.2 & 56.3    & {56.7} \\
InternVL2.5~\cite{chen2024expanding}   & 8B & 60.8  & 38.9 & 60.8 & {68.9}   & 64.2 & 60.0    & {72.0} \\
Qwen-2.5-VL~\cite{Qwen2_5VL}  & 7B   & 50.3  & 37.7 &  67.2 & {67.9}   & 66.1 & 61.3  & {68.1} \\
{Qwen2.5-VL-SFT}   & 7B   &  49.4  &  34.8 &  65.6  & 66.0   & 63.6  &  59.1  & {65.4} \\
\midrule
\multicolumn{3}{l}{\textbf{\textit{Reinforcement Fine-Tuning Models}}} \\
{Qwen2.5-VL-GRPO}   & 7B   &  50.1  & 36.7 &  66.0  & {66.2}   &  64.3 &  59.5  & {67.5} \\
Video-R1~\cite{feng2025video}  & 7B &  \underline{52.4}  &  \underline{37.1} &  63.8 &  \underline{67.7}  &  61.4  &  {57.6} &  64.8 \\
VideoChat-R1~\cite{li2025videochat} & 7B &  {52.0}  &  33.0 &  64.8 &  62.5  &  64.1   &  54.3  &  66.2  \\
VideoChat-R1.5~\cite{yan2025videochat} & 7B  &  49.6  &  36.1 &  {67.0} &  65.3  &  \underline{65.2}  &  \textbf{61.4} &  \underline{67.6} \\
VideoRFT~\cite{wang2025videorft}  & 7B &  51.1  &  {36.8} &  \textbf{68.5} &  66.6  &  59.8  & 57.0   &  62.1 \\
\rowcolor{lightblue} \textbf{TimeThink (Ours)} & 7B  &  \textbf{52.6}  &  \textbf{37.8} &  \underline{67.2} & \textbf{67.8}    & \textbf{65.5}  &  \underline{60.8}    &  \textbf{69.7} \\
\bottomrule
\end{tabular}
}
\vspace{-3mm}
\end{table*}

To comprehensively evaluate the enhancements in video understanding introduced by our reasoning constraints, we systematically assess the model's temporal localization capabilities across three benchmarks: Charades-STA, CGBench, and NExT-GQA. For the temporal grounding task on Charades-STA, the evaluation metrics (mIoU, R@0.5, and R@0.7) strictly measure the model's precision in predicting exact temporal boundaries for given language queries. For grounded Video Question Answering, CGBench utilizes metrics (mIoU, R@IoU, and A@IoU) to assess the model's segment retrieval efficacy and its ability to couple semantic answer correctness with valid temporal evidence ($tIoU>0$). Conversely, NExT-GQA emphasizes fine-grained grounding consistency and penalizes redundant predictions; its metrics (IoU, IoP, and Acc@GQA) mandate high prediction validity ($IoP\ge0.5$) alongside accurate answers, effectively evaluating the model's capacity to avoid overly conservative, long-span predictions.

{\textbf{Performance on Grounded VideoQA.}}
As demonstrated in \cref{tab:cgbench_single} and \cref{tab:nextgqa_single}, our method vastly outperforms existing SFT approaches on CGBench, surpassing InternVL2 by 1.64\% in mIoU, 3.22\% in R@IoU, and 1.37\% in A@IoU. This substantiates that TimeThink not only delivers accurate semantic answers but also exhibits exceptional temporal localization capabilities. Notably, TimeThink maintains a definitive advantage in mIoU and R@IoU, approaching the performance of GPT-4o and exceeding that of Gemini-1.5-Pro, further underscoring the critical importance of applying process rewards. On NExT-GQA, our model yields improvements of 1.8\% in mIoU, 2.9\% in mIoP, and 0.9\% in Acc@GQA compared to the Qwen2.5VL-GRPO baseline. Concurrently, it surpasses all existing Reasoning Fine-Tuning (RFT) models, establishing a new state-of-the-art among open-source models.

{\textbf{Performance on Temporal Grounding.}}
As shown in \cref{tab:charades_single}, incorporating IoU-based rewards for video clues during the intermediate reasoning process significantly enhances zero-shot grounding capabilities. On Charades-STA, our method substantially outperforms existing supervised fine-tuning (SFT) approaches in the zero-shot setting, achieving an mIoU of 55.3\%, which represents an absolute improvement of 5.1\% over the state-of-the-art VideoMind~\cite{liu2025videomind} method. Furthermore, to mitigate any potential bias arising from the data scale across training stages, we subjected our backbone model, Qwen2.5VL, to both SFT and GRPO. The empirical results demonstrate that our proposed approach significantly surpasses both the SFT and GRPO baselines in terms of R@0.5 and R@0.7, highlighting that the integration of temporal process rewards profoundly enhances the model's localization precision and visual fidelity.

\subsection{Reasoning Task Evaluation}

As demonstrated in \cref{tab:general}, our proposed method achieves state-of-the-art performance among RL-based methods on five benchmarks and secures the second-highest scores on two others. This highlights the efficiency and robust generalization capabilities of our approach across diverse video tasks. Within the reasoning domain, specifically on the VideoMMMU knowledge modeling benchmark, our model attains a performance of 52.6\%. This outperforms the Qwen2.5-VL baseline by 2.3\% and surpasses Video-R1, which relies on a substantially larger dataset of 425K instances. On the VSIBench spatial reasoning task, our model matches the performance of strong SFT baselines and outperforms alternative RL methods, confirming the preservation of spatial fidelity. 

Furthermore, on general video benchmarks, our method achieves scores of 67.8 on MLVU, 65.5 on VideoMME, and 69.7 on MVBench, establishing a new state-of-the-art among RFT Models and narrowing the performance gap with strong SFT baselines. To isolate our improvements from potential data scaling benefits, we train SFT and GRPO versions of the Qwen2.5-VL backbone. Compared to the Qwen2.5-VL GRPO model, our method yields an average improvement of 1.6\% across three reasoning tasks and 1.58\% across four general tasks. This underscores that incorporating our process-oriented reward into the reasoning trajectory not only preserves the generalization capabilities of the backbone model but also extends its proficiency in complex reasoning tasks.

\subsection{Ablation Study}
To dissect the efficacy of our proposed components, we conduct extensive ablations across multiple dimensions (\cref{tab:ablation}). We employ Qwen2.5-VL fine-tuned via SFT on the identical dataset as our baseline. The visual context length is constrained to 8K tokens, with video inputs sampled at 2 frames per second (fps) up to a maximum of 64 frames.

{\textbf{Effect of Process Reward.}}
To validate the impact of our proposed process reward, we compare the standard video GRPO configuration (a) against our approach incorporating $R_{clue}$ (b). As observed, the integration of $R_{clue}$ yields a substantial $5.0\%$ improvement on the Charades-STA temporal grounding task compared to (a). Furthermore, it improves performance on the general video benchmarks, showing gains of $0.5\%$ on VideoMME and $2.5\%$ on MLVU, as well as a $2.1\%$ increase on the VideoMMMU reasoning task. These results indicate that constraining temporal information during the reasoning process effectively enhances performance across multiple video-centric tasks, particularly those demanding precise temporal grounding.

\begin{table}[t]
\caption{\textbf{Ablation on design choices of TimeThink.}}
\vspace{-3mm}
\label{tab:ablation}
\centering
\resizebox{1.0\linewidth}{!}{
\begin{tabular}{l l cccc}
\toprule
\multirow{2}{*}{\textbf{Model}} & \multirow{2}{*}{\textbf{Factors}}  & \textbf{Charades STA} & \textbf{VideoMME} & \textbf{MLVU} & \textbf{VideoMMMU} \\
\cmidrule(lr){3-6} 
&   & mIoU & acc & acc & acc  \\

\midrule
\rowcolor{gray!15}
(-) & Qwen2.5-VL-SFT  & 48.8 & 61.4 & 59.8 & 48.1 \\

\midrule
\multicolumn{2}{l}{\textbf{\textit{Reward}}} \\
(a) & $R = R_{ans} + R_{fmt}$  & 49.3 & 63.9 & 64.2 & 49.6 \\
\rowcolor{lightblue}
(b) & $R = R_{ans} + R_{fmt} + R_{clue}$  & \textbf{54.3} & \textbf{64.4} & \textbf{66.7} & \textbf{51.7} \\

\midrule
\multicolumn{2}{l}{\textbf{\textit{Training Strategy}}} \\
(c) & Only Stage 1   & 52.4 & 63.7 & 65.6 & 50.9  \\
(d) & Only Stage 2 & 50.8 & 63.4 & 64.8 & 49.9 \\
\rowcolor{lightblue}
(e) & Stage 1 + Stage 2  & \textbf{54.3} & \textbf{64.4} & \textbf{66.7} & \textbf{51.7}  \\

\midrule
\multicolumn{2}{l}{\textbf{\textit{Process Reward Type}}} \\
(f) & Binary Reward & 51.1 & 63.7 & 65.0 & 50.5 \\
(g) & mIoU Reward & 52.9 & 64.2 & 65.9 & 51.2\\
\rowcolor{lightblue}
(h) & TimeThink & \textbf{54.3} & \textbf{64.4} & \textbf{66.7} & \textbf{51.7}\\

\midrule

\end{tabular}
}
\vspace{-5mm}
\end{table}
{\textbf{Effect of Training Strategy.}}
We evaluate several distinct training strategies. As demonstrated in \cref{tab:ablation}, training exclusively in Stage 1 (c) using only the 20K dataset with temporal clues is sufficient to achieve state-of-the-art zero-shot performance on Charades-STA. Furthermore, this strategy yields substantially greater improvements than training exclusively in Stage 2 (d) with a larger volume of conventional data. Ultimately, adopting the complete two-stage training paradigm equips the model with process-oriented temporal reasoning while fully leveraging the synergistic benefits of both stages to attain optimal performance.

{\textbf{Effect of Data Scaling.}}
As illustrated in \cref{fig:data_scaling}, we observe that integrating $R_{clue}$ during Stage 1 effectively enhances the temporal video capabilities of GRPO. Specifically, we partition the 20K dataset into subsets of 1K, 5K, 10K, and 20K instances to train Stage 1 models at varying data scales, followed by Stage 2 training on the corresponding data. The empirical results demonstrate that scaling the Stage 1 data yields consistent performance improvements across both stages. Notably, performance on temporal tasks exhibits substantial gains, while general capabilities also show measurable enhancements. This trend underscores the scalability and versatility of our proposed approach.

{\textbf{Effect of Process Reward Type.}}
To determine the optimal temporal constraint strategy, we evaluate four step-wise reward variants: (a) standard GRPO; (f) Binary Reward (1 if IoU exceeds a threshold, 0 otherwise); (g) mIoU Reward (mean IoU against all ground-truth spans); and (h) TimeThink (Max IoU). As demonstrated in the results, TimeThink (h) achieves superior performance across all metrics, whereas (f) and (g) exhibit severe vulnerabilities to reward hacking. Specifically, the Binary Reward (f) suffers from reward sparsity, lacking the fine-grained gradients necessary for precise boundary refinement. Conversely, the mIoU Reward (g) introduces target misalignment: since a single reasoning step should localize only one specific event, penalizing it for missing other ground-truths forces the model to generate excessively long, exhaustive captions to artificially maximize coverage. TimeThink's continuous Max IoU formulation intrinsically resolves these issues by accurately rewarding localized precision and effectively mitigating overly conservative predictions.

\subsection{Qualitative Results}

\begin{figure}[t]
\centering
\includegraphics[width=1.0\linewidth]{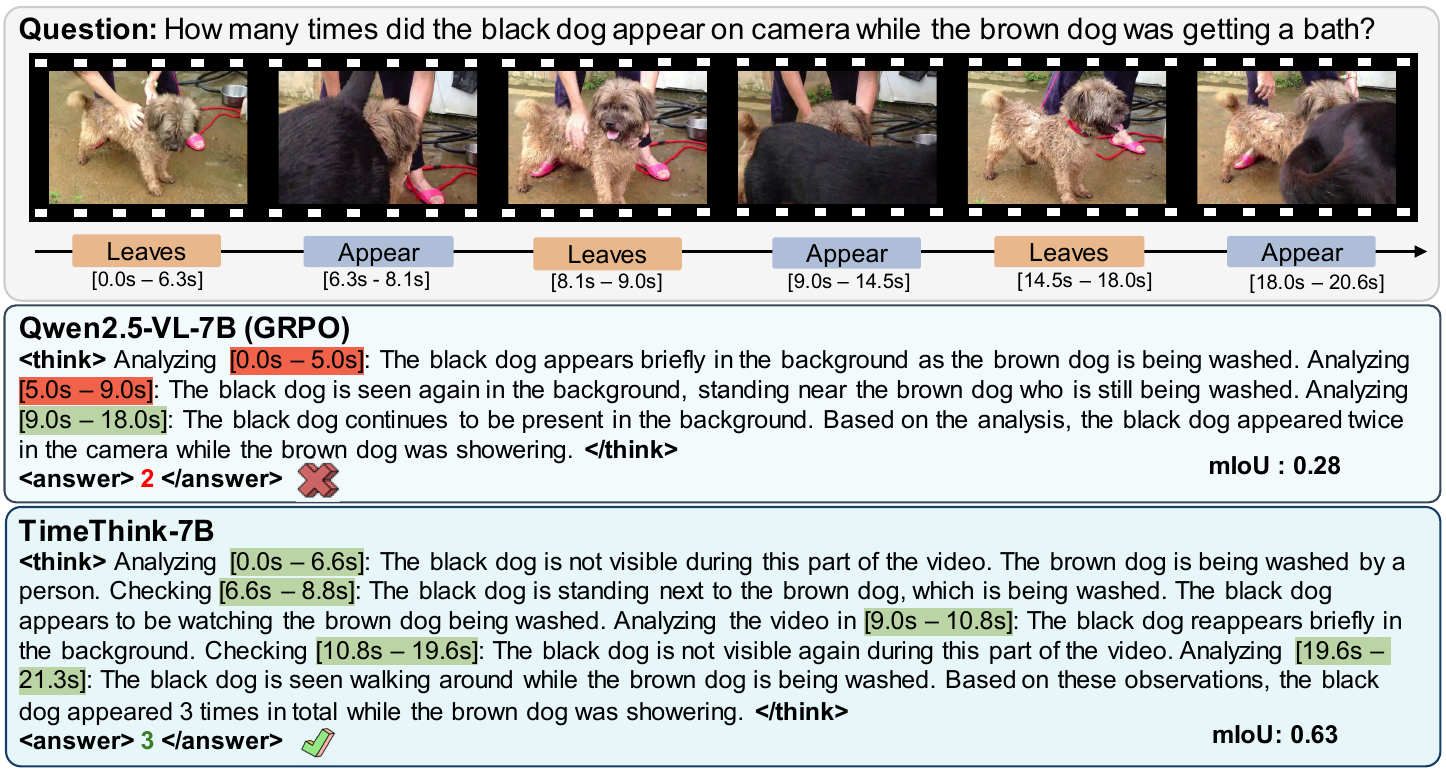}
\caption{\textbf{Qualitative comparisons on grounded QA.} The GRPO baseline yields an incorrect answer with an unfaithful reasoning trajectory. In contrast, TimeThink provides the correct answer while achieving a significantly higher mIoU for query-grounded segments, supported by a detailed and precise reasoning process.}
\label{fig:qualitive_result}
\end{figure}

To illustrate our method's efficacy, \cref{fig:qualitive_result} presents qualitative NExT-GQA examples using identical prompts and a 4 fps input. Compared to the GRPO baseline, TimeThink not only answers correctly but also generates a detailed, visually faithful reasoning trajectory with heightened sensitivity to scene transitions and action boundaries. Furthermore, TimeThink dynamically adapts to temporal resolution: evaluating at 8 fps yields noticeably more precise reasoning timestamps, whereas the standard GRPO model rigidly produces identical outputs regardless of frame rate.

\section{Conclusion}
We presented \textbf{TimeThink}, a reinforcement learning framework for temporally grounded reasoning in Video Large Language Models. By treating \emph{temporal clue steps} as fundamental reasoning units supervised via step-wise process rewards, our approach shifts optimization from final outcomes to intermediate evidence discovery. This process-oriented perspective provides a promising direction for improving reliable multimodal reasoning.

%
%
\bibliographystyle{splncs04}
\bibliography{main}
\clearpage
\setcounter{page}{1}

\makeatletter
\def\maketitlesupplementary{
   \newpage
   \begin{center}
       \Large \bfseries \savedtitle \\[0.5em]
       \large Supplementary Material \\[1.0em]
   \end{center}
}
\makeatother
   
\maketitlesupplementary

\section{Additional Ablation Studies}
\label{sec:supp_ablation}

\subsection{Impact of the Annotation Model}
To construct the TimeThink-RFT-20K dataset, we utilize an automated vision-language annotator to estimate the semantic relevance between candidate temporal clips and the question-answer pairs. In Table~\ref{tab:annotator_ablation}, we investigate how the capacity of the annotation model affects the quality of the generated process rewards and the subsequent downstream performance. 

As shown in Table~\ref{tab:annotator_ablation}, employing different annotation models for temporal clue scoring results in relatively stable downstream performance. While scaling the annotator from Qwen3-VL-8B to Qwen3-VL-235B-A22B yields slight improvements, particularly in the Charades-STA grounding task (from 53.2 to 54.3), the performance on general video QA (VideoMME) and video reasoning (VideoMMMU) remains largely consistent. This observation indicates that the effectiveness of our approach does not heavily depend on the scale or capacity of the annotation model. Instead, the primary performance gains are attributed to the TimeThink framework itself—specifically, the mechanism of incorporating step-wise temporal process rewards to guide intermediate reasoning. Even with a lightweight 8B annotator, the proposed method achieves competitive results, confirming the robustness and intrinsic value of our reasoning constraint design.

\begin{table}[!htbp]
\centering
\caption{\textbf{Impact of different annotation models for temporal clue scoring.} Performance is evaluated on VideoMME, VideoMMMU, and Charades-STA.}
\label{tab:annotator_ablation}
\begin{tabular}{lccc}
\toprule
\textbf{Annotation Model} & \textbf{VideoMME} & \textbf{VideoMMMU} & \textbf{Charades-STA} \\
\midrule
Qwen3-VL-8B & 63.9 & 51.4 & 53.2 \\
Qwen3-VL-30B-A3B & 64.1 & 51.3 & 54.0 \\
\rowcolor{lightblue}
Qwen3-VL-235B-A22B & \textbf{64.4} & \textbf{51.7} & \textbf{54.3} \\
\bottomrule
\end{tabular}
\end{table}

\subsection{Sensitivity of Process Reward Weight ($\lambda$)}
The hyperparameter $\lambda$ controls the strength of the step-wise temporal process reward relative to the global outcome reward during GRPO optimization. Table~\ref{tab:lambda_ablation} details the performance variations across different values of $\lambda$.

As demonstrated, increasing $\lambda$ places greater emphasis on the intermediate temporal reasoning process ($R_{clue}$). A higher $\lambda$ (e.g., 0.7) encourages the model to focus strictly on temporal boundary alignment, which slightly boosts the video grounding performance (Charades-STA reaches 54.5). However, higher values ($\lambda \ge 0.7$) lead to a slight degradation in general video QA (VideoMME) and complex reasoning (VideoMMMU). This implies that over-penalizing the intermediate temporal spans might restrict the model's linguistic flexibility and overfit its global semantic reasoning to strict temporal boundaries. Conversely, a lower $\lambda$ (e.g., 0.1) provides insufficient guidance for temporal evidence discovery, affecting grounding accuracy. Ultimately, we adopt $\lambda=0.5$ as the default setting to achieve a reasonable balance between accurate temporal localization and general video understanding capabilities.

\begin{table}[!htbp]
\centering
\caption{\textbf{Ablation on the process reward weight ($\lambda$).}}
\label{tab:lambda_ablation}
\begin{tabular}{lccc}
\toprule
\textbf{$\lambda$} & \textbf{VideoMME} & \textbf{VideoMMMU} & \textbf{Charades-STA} \\
\midrule
$0.1$ & 63.8 & 50.3 & 52.8 \\
$0.3$ & 64.3 & 51.5 & 53.4 \\
\rowcolor{lightblue}
$0.5$ & \textbf{64.4} & \textbf{51.7} & 54.3 \\
$0.7$ & 63.9 & 50.9 & \textbf{54.5} \\
$1.0$ & 64.0 & 51.1 & 54.3 \\
\bottomrule
\end{tabular}
\end{table}

\section{Prompt Templates}
\label{sec:prompts}
This section provides the exact prompt templates utilized during the temporal evidence data generation and the two-stage reinforcement fine-tuning process.

\subsection{Data Generation Prompt}
To automatically derive temporal evidence segments for TimeThink-RFT-20K, we prompt the teacher model to score the relevance of a candidate video clip (segmented via PySceneDetect) to a given Question-Answer pair. To prevent hallucinations and ensure accurate local evaluation, we provide the full video context alongside the specific candidate clip. The prompt template is structured as follows:

\vspace{0.5em}
\noindent\fbox{
\begin{minipage}{0.95\linewidth}
\small
\textbf{System:} You are an expert video annotator. Evaluate the semantic relevance between the candidate clip and the QA pair based on the provided Full Video Context. Follow these strict requirements: \\[0.5em]
1. Understand the Target QA: Determine the exact visual/semantic evidence required. \\
2. Evaluate the Candidate Clip: Determine if the specific clip contains this evidence. Use the Full Video Context for background, but base your score \textit{strictly} on the Candidate Video Clip. \\
3. Scoring Criteria (0 to 10 Scale): \\
\indent - \textbf{10}: Perfect Relevance. Clear, unambiguous, and complete evidence. \\
\indent - \textbf{7-9}: High Relevance. Strong evidence, might lack minor context. \\
\indent - \textbf{4-6}: Partial Relevance. Partial, ambiguous, or indirect evidence. \\
\indent - \textbf{1-3}: Low Relevance. Mostly irrelevant, tangential elements present. \\
\indent - \textbf{0}: No Relevance. Completely irrelevant. \\
4. Provide concise REASONS (1-3 sentences) explaining the score before outputting the SCORE. \\[0.5em]
\textbf{User:} \\
INPUT DATA: \\
Target QUESTION: \{question\} \\
Target ANSWER: \{answer\} \\[0.5em]
Full Video Context (For background understanding only): \\
\{full\_video\_content\} \\[0.5em]
Candidate Video Clip ([\{clip\_start\_time\} - \{clip\_end\_time\}] seconds): \\
\{clip\_content\} \\[0.5em]
STRICT OUTPUT FORMAT: \\
REASONS: <Your concise reasoning> \\
SCORE: <An integer between 0 and 10>
\end{minipage}
}
\vspace{0.5em}

\subsection{Stage 1: Temporal-Grounded RFT Prompt}
During the first stage, the model is strictly enforced to generate step-wise temporal reasoning clues enclosed in a \texttt{<think>} block.

\vspace{0.5em}
\noindent\fbox{
\begin{minipage}{0.95\linewidth}
\small
\textbf{User:} \{Question\}\\
Please think step by step. For each reasoning step, you must explicitly reference the relevant time interval in the format [start\_time - end\_time]. Enclose your entire reasoning process within <think> and </think> tags. After evaluating all necessary segments, provide the final answer within the <answer> </answer> tags.
\end{minipage}
}
\vspace{0.5em}

\subsection{Stage 2: Generalized Outcome RFT Prompt}
In the second stage, the model generalizes to broader open-ended tasks. The strict formatting constraint is slightly relaxed, allowing for more flexible reasoning while maintaining the two-block structure.

\vspace{0.5em}
\noindent\fbox{
\begin{minipage}{0.95\linewidth}
\small
\textbf{User:} \{Question\}\\
Please analyze the video and think step by step to derive the correct answer. You can reference the relevant time interval in the format [start\_time - end\_time] if necessary to support your observation. Put your reasoning process inside <think> and </think> tags. Finally, output your concise conclusion inside <answer> and </answer> tags.
\end{minipage}
}
\vspace{0.5em}

\section{Implementation Details}
\subsection{Training Configuration}\label{sec:sup_detail}
Table~\ref{tab:hyperparams} provides a comprehensive summary of the hyperparameters employed in the training of TimeThink. We directly apply Group Relative Policy Optimization (GRPO) on the non-reasoning Qwen2.5-VL-7B backbone.

\begin{table}[!htbp]
\centering
\caption{\textbf{Hyperparameter configuration for TimeThink training.} The model undergoes a two-stage reinforcement fine-tuning protocol.}
\label{tab:hyperparams}
\begin{tabular}{lc}
\toprule
\textbf{Hyperparameter} & \textbf{Value} \\
\midrule
\multicolumn{2}{l}{\textit{Model Configuration}} \\
Backbone & Qwen2.5-VL-7B \\
Max Visual Sequence Length & 24,576 \\
Max Context Length & 32,768 \\
\midrule
\multicolumn{2}{l}{\textit{Training Optimization}} \\
Global Batch Size & 64 \\
GRPO Group Size & 8 \\
Learning Rate (lr) & $1 \times 10^{-6}$ \\
Optimizer & AdamW \\
LR Schedule & 1e-6 \\
Weight Decay & 0.05 \\
\midrule
\multicolumn{2}{l}{\textit{Data \& Hardware}} \\
Input FPS & 2 \\
Compute Resources & 32 $\times$ NVIDIA H100 \\
Training Time (Stage 1) & $\approx$ 1 Day \\
Training Time (Stage 2) & $\approx$ 3 Days \\
\bottomrule
\end{tabular}
\end{table}

Our training methodology initiates directly with a two-stage GRPO pipeline, bypassing the conventional Supervised Fine-Tuning (SFT) cold-start. During the first stage, we utilize the TimeThink-RFT-20K dataset to train the model to ground its intermediate reasoning trajectory. In the second stage, we transition to the LLaVA-Video-178K dataset for generalized video comprehension. We also augment this distribution with 77K timestamp-grounded samples from DiDeMo and ActivityNet Captions to enhance the extraction of key temporal information.

\section{Evaluation Settings}
We evaluate TimeThink across a suite of video-language benchmarks: 1) Video Reasoning Benchmarks, including VideoMMMU, VSIBench, and MMVU; 2) General Video Understanding, comprising MLVU, VideoMME, Long VideoBench, and MVBench; and 3) Temporal Video Benchmarks, utilizing Charades-STA, CGBench, and NEXT-GQA. 

Table~\ref{tab:eval_settings} details the frame sampling configurations. All quantitative assessments are conducted using the unified evaluation protocols provided by the lmms-eval framework to ensure standardized evaluations.

\begin{table}[!htbp]
\centering
\caption{\textbf{Evaluation settings summary for each benchmark.} \textbf{FPS} denotes the sampling frames per second, and \textbf{\# F} represents the maximum number of sampling frames allowed.}
\label{tab:eval_settings}
\begin{tabular}{lcc}
\toprule
\textbf{Benchmark} & \textbf{FPS} & \textbf{\# F (Max Frames)} \\
\midrule
\multicolumn{3}{l}{\textit{Video Reasoning}} \\
VideoMMMU & 1 & 512 \\
VSIBench & 2 & 256 \\
MMVU & 1 & 512 \\
\midrule
\multicolumn{3}{l}{\textit{General Video Understanding}} \\
MLVU & 1 & 512 \\
VideoMME & 2 & 256 \\
Long VideoBench & 1 & 512 \\
MVBench & 2 & 128 \\
\midrule
\multicolumn{3}{l}{\textit{Temporal Video}} \\
Charades-STA & 4 & 128 \\
CGBench & 2 & 512 \\
NEXT-GQA & 2 & 256 \\
\bottomrule
\end{tabular}
\end{table}

\section{Additional Samples}
\label{sec:additional_qualitative}

\begin{figure}[ht]
    \centering
    \includegraphics[width=1.0\linewidth]{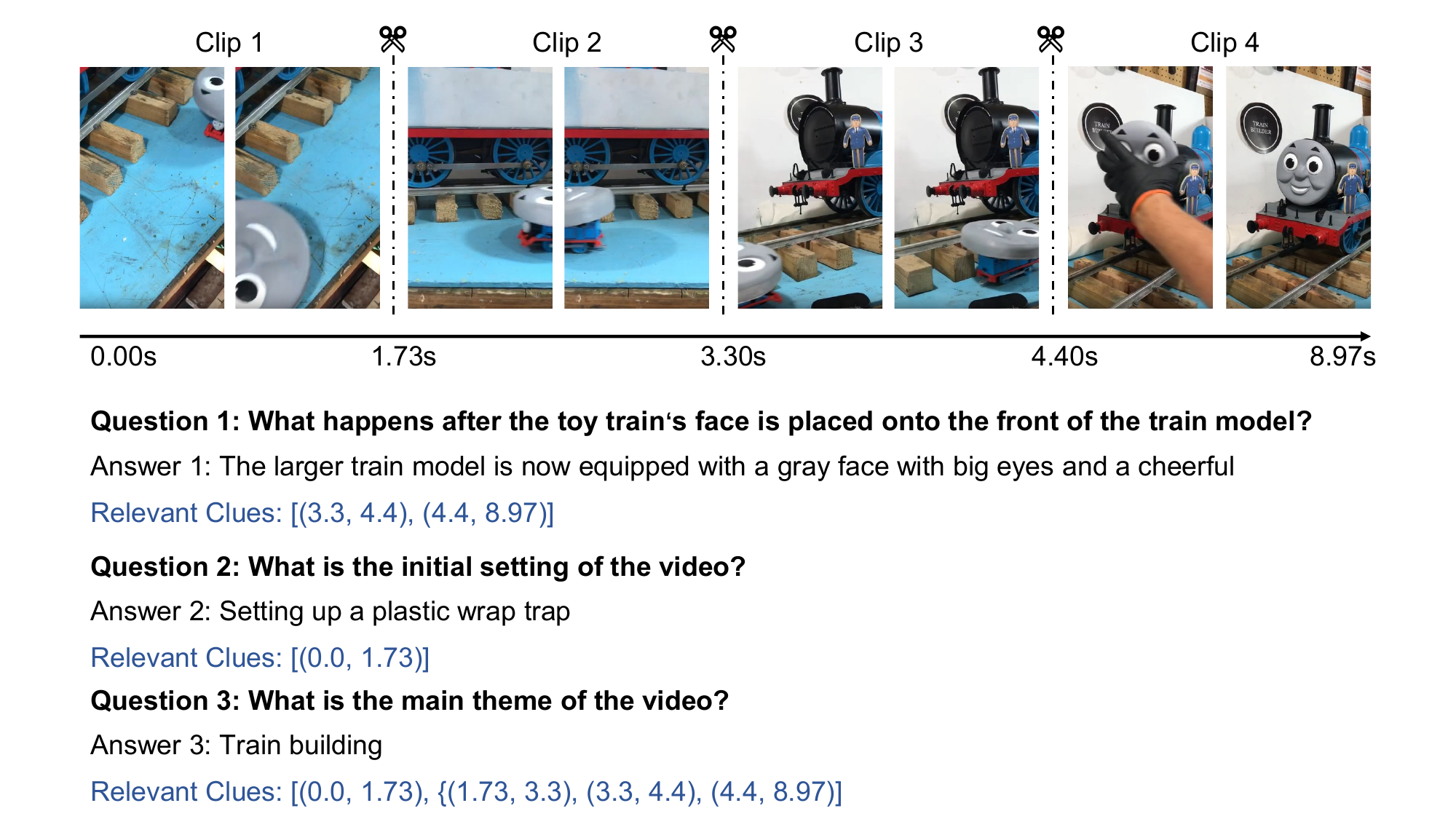}
    
    \vspace{1em} 
    
    \includegraphics[width=1.0\linewidth]{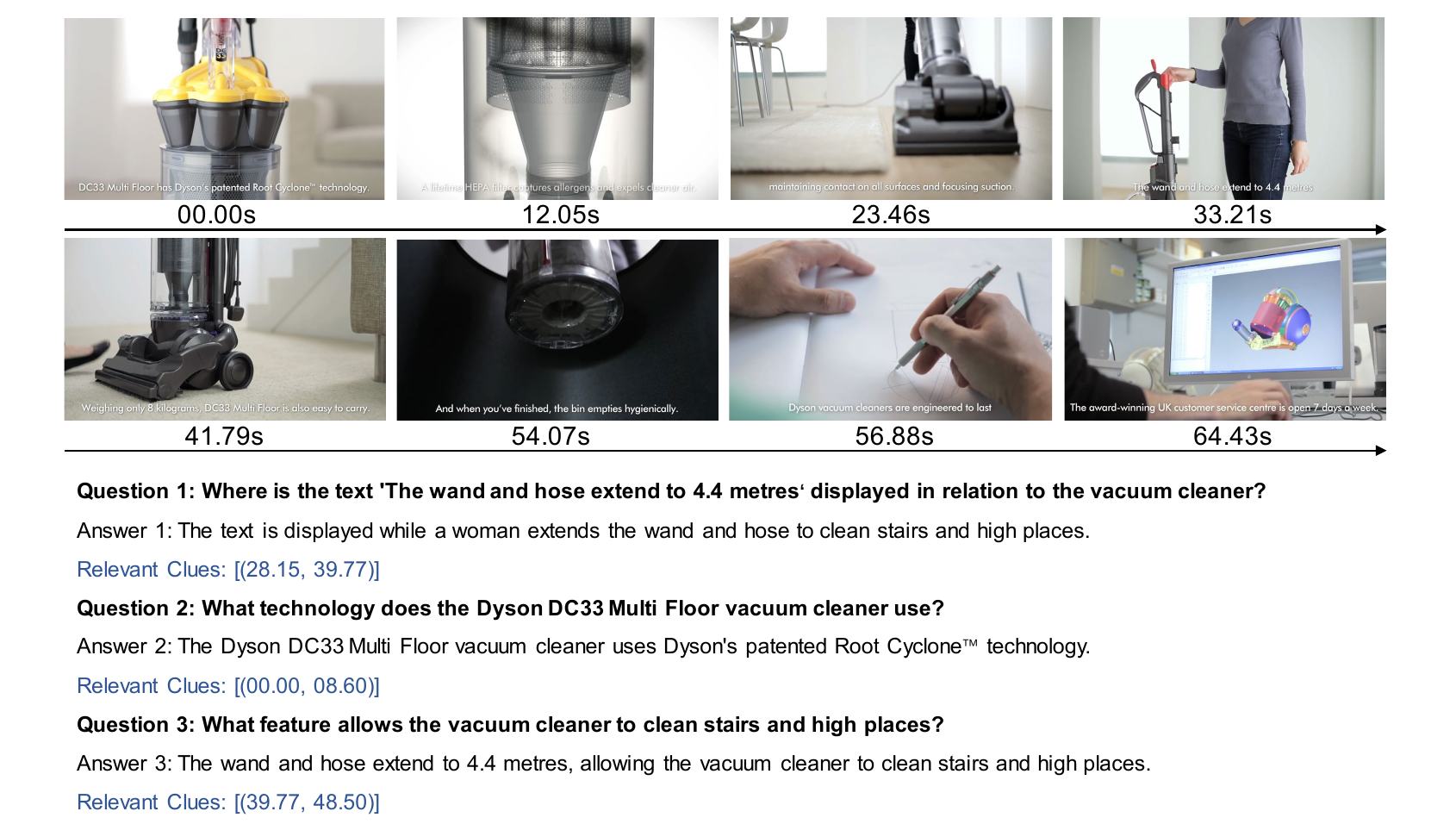} 
    
    \caption{\textbf{Exemplars from the TimeThink-RFT-20K dataset.} Each video sequence is systematically segmented into multiple candidate clips. Distinct question-answer pairs are explicitly grounded with their respective, contextually relevant temporal clues.}
    \label{fig:dataset_exemplars}
\end{figure}

As illustrated in Figure~\ref{fig:dataset_exemplars}, the TimeThink-RFT-20K dataset provides precise, query-aware temporal grounding. For instance, in a video depicting the assembly of a toy train, the continuous sequence is partitioned into four distinct temporal clips. A query regarding the ``initial setting'' explicitly anchors its evidence in the first clip (0.00s - 1.73s). Conversely, a query addressing a subsequent action---such as events occurring after the train's face is attached---identifies relevant clues exclusively within the latter segments. Furthermore, a holistic query concerning the ``main theme'' correctly aggregates evidence across the entire video timeline. This exemplar demonstrates how our dataset supplies fine-grained, question-specific temporal supervision, compelling the model to dynamically align its reasoning process with the appropriate temporal intervals based on the specific semantic context of the query.

\section{Limitations}
Our model demonstrates strong performance and temporal coherence on open-domain tasks. However, a specific limitation emerges when processing hour-long videos using frames sampled with large time intervals. Under these extreme long-context conditions, executing dense captioning forces the model to comprehensively traverse the entire extensive video sequence. This exhaustive temporal grounding behavior consequently results in excessively long text outputs, which may impact reading efficiency and inference speed.

\end{document}